\documentclass[a4paper,man,natbib,floatsintext]{apa6}
\usepackage{apacite}
\usepackage{subfig}
\usepackage[english]{babel}
\usepackage[utf8x]{inputenc}
\usepackage{amsmath}
\usepackage{graphicx}
\usepackage[colorinlistoftodos]{todonotes}
\usepackage{diagbox}
\usepackage{indentfirst}
\usepackage{adjustbox}
\usepackage{xcolor}
\usepackage{multirow}
\usepackage{caption}
\usepackage[pagewise]{lineno}

\usepackage{setspace}

\begin{document}

\raggedbottom

\captionsetup[table]{
  labelsep = newline,
  justification=raggedleft,
  singlelinecheck=off,
  labelsep=colon,
  skip = \medskipamount}

\begin{titlepage}

\begin{center}

\large Design a Sustainable Micro-mobility Future: Trends and Challenges in the US and EU
\\ 

\normalsize

\vspace{25pt}
Lilit Avetisyan\\ 
Industrial and Manufacturing Systems Engineering, University of Michigan-Dearborn\\
\vspace{15pt}
Chengxin Zhang\\ 
Industrial and Manufacturing Systems Engineering, University of Michigan-Dearborn\\
Sue Bai\\ 
Honda Research Institute USA, Inc., Industrial and Manufacturing Systems Engineering, University of Michigan-Dearborn\\
\vspace{15pt}
Ehsan Moradi Pari\\
Honda Research Institute USA, Inc. \\
\vspace{15pt}
Fred Feng\\ 
Industrial and Manufacturing Systems Engineering, University of Michigan-Dearborn\\
\vspace{15pt}
Shan Bao\\ 
Industrial and Manufacturing Systems Engineering, University of Michigan-Dearborn, University of Michigan Transportation Research Institute\\
\vspace{15pt}
Feng Zhou\\ 
Industrial and Manufacturing Systems Engineering, University of Michigan-Dearborn\\
\vspace{15pt}

\end{center}
%





\end{titlepage}
\shorttitle{}

\section{ABSTRACT}

Micro-mobility is promising to contribute to sustainable cities in the future with its efficiency and low cost. To better design such a sustainable future, it is necessary to understand the trends and challenges. Thus, we examined people’s opinions on micro-mobility in the US and the EU using Tweets. We used topic modeling based on advanced natural language processing techniques and categorized the data into seven topics: promotion and service, mobility, technical features, acceptance, recreation, infrastructure and regulations. Furthermore, using sentiment analysis, we investigated people’s positive and negative attitudes towards specific aspects of these topics and compared the patterns of the trends and challenges in the US and the EU. We found that 1) promotion and service included the majority of Twitter discussions in the both regions, 2) the EU had more positive opinions than the US, 3) micro-mobility devices were more widely used for utilitarian mobility and recreational purposes in the EU than in the US, and 4) compared to the EU, people in the US had many more concerns related to infrastructure and regulation issues. These findings help us understand the trends and challenges and prioritize different aspects in micro-mobility to improve their safety and experience across the two areas for designing a more sustainable micro-mobility future.


\textbf{Keywords:} Micro-mobility, Social media, Sustainability, Natural Language Processing.


\newpage
\section{INTRODUCTION}
The growth of transportation has raised the need for compact, flexible, and more sustainable forms of transportation. Recent developments in the micro-mobility industry show that these devices might address this issue and offer people safer and cheaper trips with reduced travel time. According to the Society of Automotive Engineers (SAE) definition \citep{sae}, micro-mobility refers to a range of small, less than 500 pounds (227 kg) lightweight, fully motorized or motor-assisted devices operating at a speed below 30 mph (48 km/h) and ideal for trips up to 10 km. Typical examples include e-bikes, e-scooters, e-unicycles and e-skateboards, and some of them are widely used as personal or shared transportation devices \citep{price2021micromobility}. 
The global micro-mobility market has been increasing over the years.  According to the NACTO \citep{nacto2020}, 136 million trips were generated by shared micro-mobility in 2019 in the U.S., which was 60\% more than 2018. Thus, micro-mobility devices can be well integrated into the overall urban design process of smart and sustainable transportation in the near future. With the sustainable design and development goal, we should not only consider technical challenges and requirements (e.g.,  battery and material), but also complement and constrain the design and development process by social, infrastructural, and political schemes for a sustainable future \citep{jiao2022newdesign}. 
Hence, it is important to investigate such factors that influence the trend of and the challenges associated with micro-mobility so far.    

Micro-mobility device have been shown to improve the overall sustainability goal in many urban areas around the world thanks to its various benefits. The light weight and small size of those devices provide options for short distance as well as “last mile” travels \citep{gossling2020integrating}. Micro-mobility devices are popular among riders because they are convenient and efficient; some customers express that they can “ride at higher speed with less effort”, especially in hilly areas \citep{ling2017differences}. People also use those devices for recreation and exercise purposes, and feel relaxed and fun while riding \citep{haustein2016bike}. The spread of micro-mobility sharing programs reduces the cost of short-distance travel and makes them a cost-effective mode of transportation \citep{tark2020patterns}. It is also beneficial from a business point of view because the design and development cost of micro-mobility devices is considerably lower compared to cars. 
In many metropolitan areas, micro-mobility plays an important role for urban sustainability by providing solutions for traffic congestion and reducing carbon emissions \citep{abduljabbar2021role}. Since those devices do not depend on fossil fuels, they are more sustainable compared to motor vehicles.

However, in order to include micro-mobility devices in the sustainable design process in the future urban areas, it is important to investigate the challenges associated with them nowadays. Abduljabbar et al. \citeyearpar{abduljabbar2021role} showed that the key barriers were vandalism and theft, redistribution, availability, safety concerns, and relevant regulations dealing with those issues.
Pimentel and Lowry \citeyearpar{pimentel2020taming} examined micro-mobility related laws in the US and pointed out that the improper parking of bicycles, e-bikes, or e-scooters in sidewalks or other shared paths was one of the key barriers about shared micro-mobility systems. Pimentel and Lowry \citeyearpar{pimentel2020taming} also pointed out that one of the potential challenges for shared micro-mobility was the lack of legal infrastructure, the unclear regulation and design of riding and parking areas may reduce the motivation of potential users. 
In terms of safety,  \citet{kroyer2021bicycles} conducted a systematic review of transportation safety of micro-mobility, they found that single accidents not involving other vehicles accounted for the majority of the accidents. Crashes with motor vehicles accounted less than one third of the total accidents, but they led to most of the fatal accidents involved micro-mobility. Moreover, the authors also found that many safety investigations regarding micro-mobility reported a low usage of helmet, which were more vulnerable to infrastructure defects, such as potholes and ditches.
U.S. Consumer Product Safety Commission \citeyearpar{tark2020patterns} summarized the statistics of injuries and crash patterns of micro-mobility from 2017 to 2019, and they concluded that the leading causes of micro-mobility involved injuries were unspecified falls, accompanied with other notable hazards, such as loss of controls, collision with motor vehicles, and road surface quality issues. Thus, a more systematic approach to addressing the safety issues associated with micro-mobility might be more effective by integrating vehicle and infrastructure design to the overall socio-technical system in the urban areas.

Previous studies examined literature and/or related laws about the benefits and challenges associated with the design and development of micro-mobility. In this study, we aim to examine the self-reported data of e-bike and e-scooter (the major forms of micro-mobility) users through social media across the US and the EU (i.e., tweets collected in Italy, Germany, France, Estonia, Belgium, Croatia, Ireland, Malta, The Netherlands, Poland, Portugal, Sweden, Finland, Czech Republic,	Bulgaria, Luxembourg, Austria, Denmark, Spain, Romania, Latvia, Lithuania, Greece, Slovenia, Hungary, Cyprus, Slovakia, and  United Kingdom) for comparisons.  We used social media platforms as a data source since previous studies \citep{zhou2015latent, drus2019soc_med,zhou2020machine} showed that such data provided valuable insights into customer needs with various products and determine key aspects of their interests for redesign and improvement. Then, we applied natural language processing (NLP) techniques based on the state-of-the-art deep learning models. Such models have proved to be effective in transportation \citep{zhang2022disengagement,ayoub2022cause} to uncover insights.

In summary, the contributions of this study are: 
\begin{itemize}
    \item We collected and analyzed data from Twitter and developed a taxonomy of main topics for micro-mobility in the US and the EU.
    
    \item We investigated the sentiment of collected data in both regions to understand users' attitudes regarding different aspects of micro-mobility devices and services.
    
    \item We provided insights into the similarities and differences of the US and the EU for regional comparisons, including predominant causes of particular sentiments and attitudes in the two regions to provide recommendations for designing a sustainable micro-mobility future.
 
\end{itemize}

\section{Related Work}
\subsection{E-bikes}
A limited number of studies have examined the development and challenges of e-bike, such as adoption, safety, infrastructure. \citet{ling2017differences} investigated usage and expectation differences between e-bike and conventional bike owners in the USA and pointed out that financial issues were the main hindrance affecting e-bike purchases, such as the purchase prices cycling distance per battery charge, and maintenance costs, followed by availability concerns that no e-bikes or no desire modes of e-bikes were available in the living areas. 
\citet{macarthur2018north} conducted an online survey to explore that motivation for people to buy or use e-bikes. The findings showed three main issues: 1) required services for e-bikes (e.g., the electric and battery issues), 2) the concerns of the range of the battery, and 3) safety concerns that about one third of the users had experienced crashes and about one tenth of the users believed e-bikes had contributed to their crashes. Other issues included 1) the weight of e-bikes, 2) the difficulty in riding under inclement weather, and 3) the cost and relatively short travel distance. 
\citet{zagorskas2019challenges}
investigated the issues accompanied with increased use of e-powered personal mobility in Europe cities and found that most of the micro-mobility devices (i.e., e-bikes, e-mopeds, and e-scooters) usually have small wheels to be fordable and compact. However, those small wheels were very sensitive to road surfaces. Riding these small devices on uneven pavements was not smooth, causing the discomfort and risks of falling.

\subsection{E-scooters}
E-scooters are another emerging mode of the micro-mobility devices, and previous studies explored user satisfaction and experience, safety concerns, and public acceptance. For example, \citet{aman2021listen} analyzed about 12,026 reviews from the Lime and Bird App store and found that safety issues significantly influenced satisfaction of the e-scooter users, such as the lack of regulations on speed and helmet, lane usage, and minimum age to ride. Other factors that influenced user experience were also identified, including quality of the battery, especially the charge level, pricing and payment of shared rides, customer service and refund, maps in finding available e-scooters. 
\citet{gossling2020integrating} studied on the internet media reports from ten major cities in Europe and the US, and they also found that the expensive cost of battery replacement was one of the major barriers for users. 
\citet{kopplin2021consumer} conducted an e-scooter survey through Facebook and collected data from 749 responses, and their result showed that e-scooter users expressed negative feedback on e-scooter damage and pollution that caused by the random disposal of e-scooters. 
\citet{dahl2020scooter} investigated the secondary data of e-scooters based on previous studies and indicated that the short lifespan, improper parking, and concerns about potential crash were the main barriers of e-scooter usage. 
Compared to the safety issues of e-bikes, besides the single accidents, collision with vehicles and visibility condition seem to have more impact on e-scooter related injuries. \citet{yang2020safety} analyzed the e-scooter crashes from 169 news reports from 2017 to 2019, the news crashes showed that colliding with vehicles and falling-off were the two leading causes of e-scooter involved crashes. They also stated that the accident rate at night was high, which related to collision with inanimate objects under low visibility. Blomberg et al. \citeyearpar{blomberg2019injury} conducted a survey study through emergency medical departments by collecting information from 468 scooter injuries from 2016 to 2019, including 112 e-scooters related injuries. They found that the most frequent injure type was falling off from the e-scooters, which accounted for about 86.6\% of total e-scooter incidents, followed by collision with vehicles or moving objects (8.9\%) and collision with objects (4.5\%). The result also showed that about one third of the e-scooter accidents happened at nighttime (between 23:00 and 07:00 hours). Despite the fact that the survey's findings in the EU indicated that e-scooters were more frequently utilized for shopping and entertainment than for commuting, some EU countries (e.g., France and Germany) implemented minimum speed and age restrictions on e-scooters in order to facilitate a quick and smooth integration in public transportation \citep{raptopoulou2020first}. 
Useche et al.\citeyearpar{useche2022environmentally} conducted a systematic review of US and EU studies and pointed out that poor risk perception and lack of regulations were the main factors affecting e-scooter riders' safety. Furthermore, they indicated that most of the e-scooter riders were young, male and highly educated people living in urban areas, and used the devices for short distances and leisure purposes.
\citet{campisi2021gender} investigated the use of e-scooters among females in Italy and identified a gender equality issue in micro-mobility since the perceived safety and comfort were low, and concluded that these results should be taken into consideration when designing micro-mobility schemes (e.g., infrastructure, devices). 
\subsection{Micro-mobility Sharing Programs}
Previous researchers also examined micro-mobility sharing programs and other electric micro personal devices. \citet{abduljabbar2021role} found that the damage and theft were one of the issues for bike-sharing programs, better maintenance and lock protection were needed. Moreover, information sharing about availability, careless parking, and disposal issues caused public concerns as well.  While \citet{tuncer2020scooters} reported that rental e-scooters intensified the conflict with public space, user experience was important to attract more users as an enjoyable way of hacking the city. 
\citet{sunio2020social} presented a bike-sharing program with a more explicit social enterprise model to achieve a broad impact for sustainability.  
\citet{boglietti2021survey} reviewed 90 studies between 2014 and 2020 and identified issues associated with the spread of e-powered micro-mobility devices, such as e-scooter, e-kick scooters and self-balancing devices. The results indicated that the lack of legislation on speed or other requirements, spatial issues with other road users, and related economic concerns prevented the diffusion of e-powered micro-mobility devices to a large extent. \citet{loustric2020} studied about  city propensity of adopting micro-electric devices, including e-bikes, scooters, hover-boards, and micro electric devices. The authors pointed out that the major consumer challenge for micro-mobility was their limited power and the limitation on travel distance compared to other motor vehicles. \citet{handy2022} believed that increasing awareness of e-bikes would increase their adoption. The influence on considering commuting by e-bikes was not significant, even though the experiment of installing an e-bike sharing system in smart phones increased awareness three times. They suggested that the increase of charging stations and numbers of available e-bikes should change the results.

\section{METHOD}

\begin{figure}[bt]
\centering
\includegraphics[width=1\textwidth]{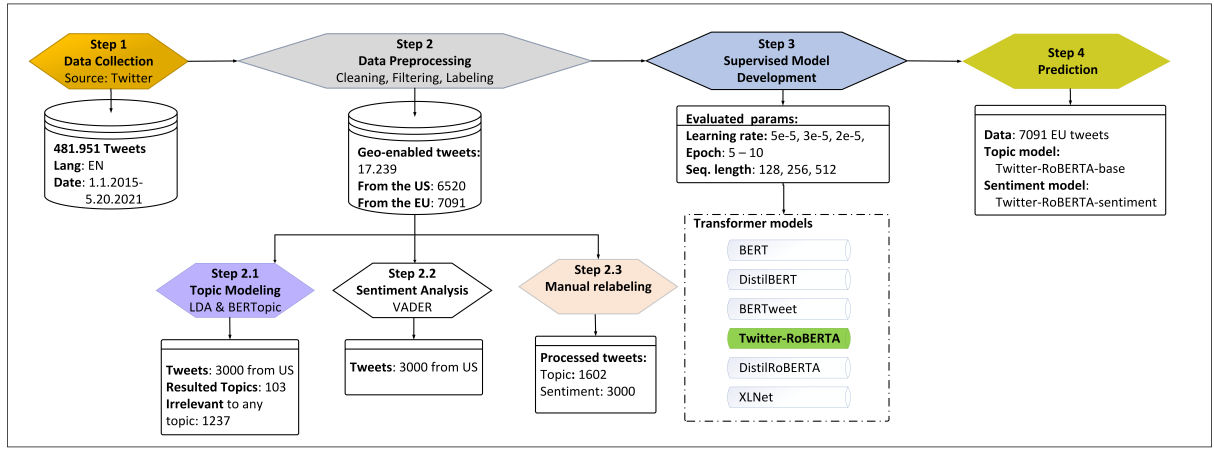}
\caption{Study Procedure.}
\label{fig:process}
\end{figure}

The overall process of the method is shown in Figure \ref{fig:process} and we describe each section below in details.

\subsection{Data Collection}
In this study, we mined and analyzed Twitter data that were related to micro-mobility in the USA and the EU. Twitter.com is one of the biggest media platforms, where users can post (tweet) their opinions and experience with micro-mobility and shared systems, as well as for manufacturers to keep the audience up to date with new updates and industry trends. Moreover, Twitter grants programmatic access to their huge public database, enabling data collection and analysis in details. Through the Twitter API (Application Programming Interface), we identified and extracted a total of 481,951 tweets in English, containing at least one of the micro-mobility keywords (i.e., e-bike, e-scooter, micro-mobility) from January 2015 to May 2021. Since our targets were the USA and the EU, we applied additional location filtering, which resulted in a total of 13,611 tweets (i.e, US: 6,520 and EU: 7,091).
\subsection{Data Preprocessing}
The collected data was processed by text analysis 
using NLP, aiming to uncover insights from unstructured data.
There are two main text analysis techniques used in this study: 1) Topic modeling is a process of clustering text data based on similar characteristics of individual tweets, and classifying them relevant to their associations to particular topics, and 2) sentiment analysis or opinion mining is a process to determine the polarity of the text data (i.e., positive, neutral, or negative). 
The following two sections describe in details the implementations of these methods on our data set.

\subsubsection{Topic modeling}

In order to extract the topics in the collected tweet data, we used two unsupervised learning techniques for topic modeling, including LDA (Latent Dirichlet Allocation) \citep{blei2003LDA} and BERTopic \citep{grootendorst2020bertopic} 
The LDA model is a Bayesian probabilistic three-level hierarchical model, which maps texts to different topics followed by mapping of separate words in the text to that topics. This model is a kind of "bag-of-words", method where the order of words is not considered.
BERTopic was built on BERT (Bidirectional Encoder Representations from Transformers) \citep{devlin2019BERT} embeddings for a document, where BERT was a transformer-based machine learning model, pre-trained on large corpora for different types of NLP tasks. 

First, BERTopic used the sentence-transformers package in Pytorch with 512 dimensions. In order to balance the computational resources and performance, Distilbert was used instead as the model to transform raw text data into sentence embeddings in Python in Google Colab. The embeddings at the sentence level worked much better than word level encoding. Second, once the sentence embeddings were obtained, their dimensions needed to be reduced due to the fact that clustering was often performed poorly in high dimensions due to the so-called curse of dimensionality, where all distances were large in sufficiently high dimensional space. Uniform Manifold Approximation and Projection (UMAP) \citep{2018arXivUMAP} was used in BERTopic to search a low dimensional projection of the embeddings that had the closest possible equivalent fuzzy topological structure. The model reduced the dimensionality to 5 while keeping the size of the local neighborhood at 15. Third, the model used the HDBSCAN (Hierarchical Density-Based Spatial Clustering of Applications with Noise) \citep{mcinnes2017hdbscan} algorithm to cluster similar embeddings together after dimension reduction. HDBSCAN was able to cluster the embeddings meaningfully with minimal parameter tuning. It did not need to specify the number of clusters (unlike k-means) or thresholds to be in a neighborhood and it could identify noisy members that were far away from the neighbors. However, it might end up with a large number of embeddings without a specific cluster. Fourth, after the documents were clustered, the model applied class-based TF-IDF scores to find the most important words within each cluster to represent the topics. The model might produce a large number of granular clusters. In order to merge possible similar topics together, we can increase one parameter, i.e., the minimal number of members within a cluster, to produce a smaller number of clusters based on the previous results. Thus, Step 3 and Step 4 can be iterated in order to tune the results. 

We used both BERTopic and LDA to identify topics from the US tweets. During this process, we tuned the results with different granularity levels in order to determine the appropriate number of topics. Then, within each topic, we manually read the most relevant tweets to interpret the results and assigned a topic name as the label of the cluster. The BERTopic model performed much better than the LDA model. The overall process was efficient and effective. However, in order to further improve the accuracy of the topic modeling, we manually checked all the labels generated by the BERTopic model with four human raters independently with 98.75\% agreement. In such a way, we have a set of labeled topics that were used to train a supervised topic models (see Twitter-roBERTa below) to predict the labels for the EU tweets.

\subsubsection{Sentiment analysis}
We performed sentiment analysis using the VADER (Valence Aware Dictionary and sEntiment Reasoner)\citep{hutto2014vader} lexicon and rule-based sentiment analysis tool. Unlike machine learning methods, the VADER approach was especially built for social media data with an emotion dictionary and with a set of five heuristics (i.e., punctuation, capitalization, intensifiers, conjunctions, negation), which determined the sentiment polarity of sentences and an intensity score at the same time.  Based on the output of the VADER score, we then manually checked and relabeled 3,000 randomly selected US tweets. In order to improve the reliability of the labels, the data was relabeled by four human raters independently. Furthermore, additional discussion sessions were involved to resolve inconsistent labels with an inter-rater reliability of 97.6\%.

\subsection{Supervised Model Development and Prediction}

We further fine-tuned sentiment and topic prediction using the labeled data obtained previously. A total of six transformer-based ML models (i.e. BERT, BERTweet \citep{nguyen2020bertweet}, DistilBert \citep{sanh2019distilbert}, DistilRoberta \citep{sanh2019distilbert}, Twitter-roBERTa \citep{barbieri2020tweeteval} and XLNet \citep{yang2019xlnet}) were tested and used for comparison.
Transformers were first introduced in 2017 by Google researchers \citep{vaswani2017attention} and became turning point in NLP by processing large scale of documents with high performance and accuracy. 
Transformer’s architecture includes two main segments called encoder and decoder that process the input and output sequences respectively and generate one item at the time.
First, the encoder takes the word sequence as input and chooses on what tokens the model should pay “attention” to learn and predict the next word. Second, decoder begins with a start token and takes a list of previous outputs as inputs, as well as the encoder outputs that contain the attention information from the input, applies positional encoding to determine the right positions of the word in the sequence, and stops decoding when it generates a token as an output. 
During the model evaluation, different hyperparameters were tested (i.e., maximum sequence length (128, 256, 512), epochs (5-10) and learning rate ( $5\times 10^{-5}, 3\times 10^{-5}, 2\times 10^{-5}$)), aiming to find the best training parameters for our data set and fine-tune the models. 

Results showed that the Twitter-RoBERTa model had the highest f1 performance score (see Table \ref{table:perf}) both for topic and sentiment prediction, and was selected for further prediction for the EU tweets. This model was developed on RoBERTa model and was trained on ~58M tweets to evaluate Twitter-specific data. The RoBERTa model uses the BERT’s strategy of masked language modeling and prediction with different hyperparameters, but excludes the next sentence prediction task included in the BERT architecture using full sentences or documents as an input for training data. The model was trained on 160GB language corpora with larger batches. 
To improve the performance, RoBERTa also switched to the dynamic data masking technique, which encodes the input data 10 times, each time using different encoding pattern to avoid masking the same words in the sequence.  This allows RoBERTa to achieve better performance in language modeling tasks compared to BERT. \citet{barbieri2020tweeteval} developed the Twitter-RoBERTa model by post-training the general RoBERTa model with a large Twitter corpus.

Finally, we reviewed the negative and positive tweets per topic for the purpose of finding out the main causes of the complaints and the factors that attracted people in micro-mobility. The aim of the detailed investigation was to point out the vulnerable areas that might influence the progress of micro-mobility development and improvement in the future. 

\subsection{Comparisons Between US and EU}
Two comparison processes were applied to understand the differences between two regions with statistical analysis. First, for each region, we identified the usage patterns, main trends, and challenges under each topic, then we compared the results to identify the regional differences. Second, we performed Chi-squared tests to analyse the distributions and associations of positive, negative, and neutral tweets across the seven topics in each region. Finally, we performed Chi-squared tests to identify the statistical differences of regions regarding sentiment and topic distributions.

\section{RESULTS}
\subsection{Micro-mobility topics}
\subsubsection{US results} In the topic modeling process, the LDA and BERTopic unsupervised models were performed on the 3,000 US tweets. The results of manual labeling showed that BERTopic performed better than LDA. The BERTopic model resulted in 103 fine-grained topics that share common characteristics with similarity value above 0.915, and 1,237 tweets did not belong to any of the topics. Furthermore, we manually evaluated and relabeled the results and removed addition 161 tweets because of irrelevant content. The manual re-grouping at the lower granularity level of the remaining 1,602 tweets resulted in a total of 17 topics (i.e., Advertising, Recreation, Transportation, Infrastructure, New Release, New Customer, Mobility, Acceptance, Financing, Customer Service, Safety, Regulations, Body features, Battery, Speed, Power, and Sharing). However, the visual map of inter-topic distance indicated that majority of the topic clusters were fairly close to one another (e.g., body features, speed, power and battery were mapped on top of each other) and several topics contained only a small number of tweets (e.g., customer service: 9, sharing: 12) comparing to others, which negatively influenced the performance of supervised models. To resolve this issue,  we merged topics with a small numbers considering the distances among them. We finally identified seven topics, including 1) Promotion and service, 2) Mobility, 3) Technical features, 4) Acceptance, 5) Recreation, 6) Infrastructure, and 7) Regulations.

\begin{figure}[bt]
\centering
\includegraphics[width=1\textwidth]{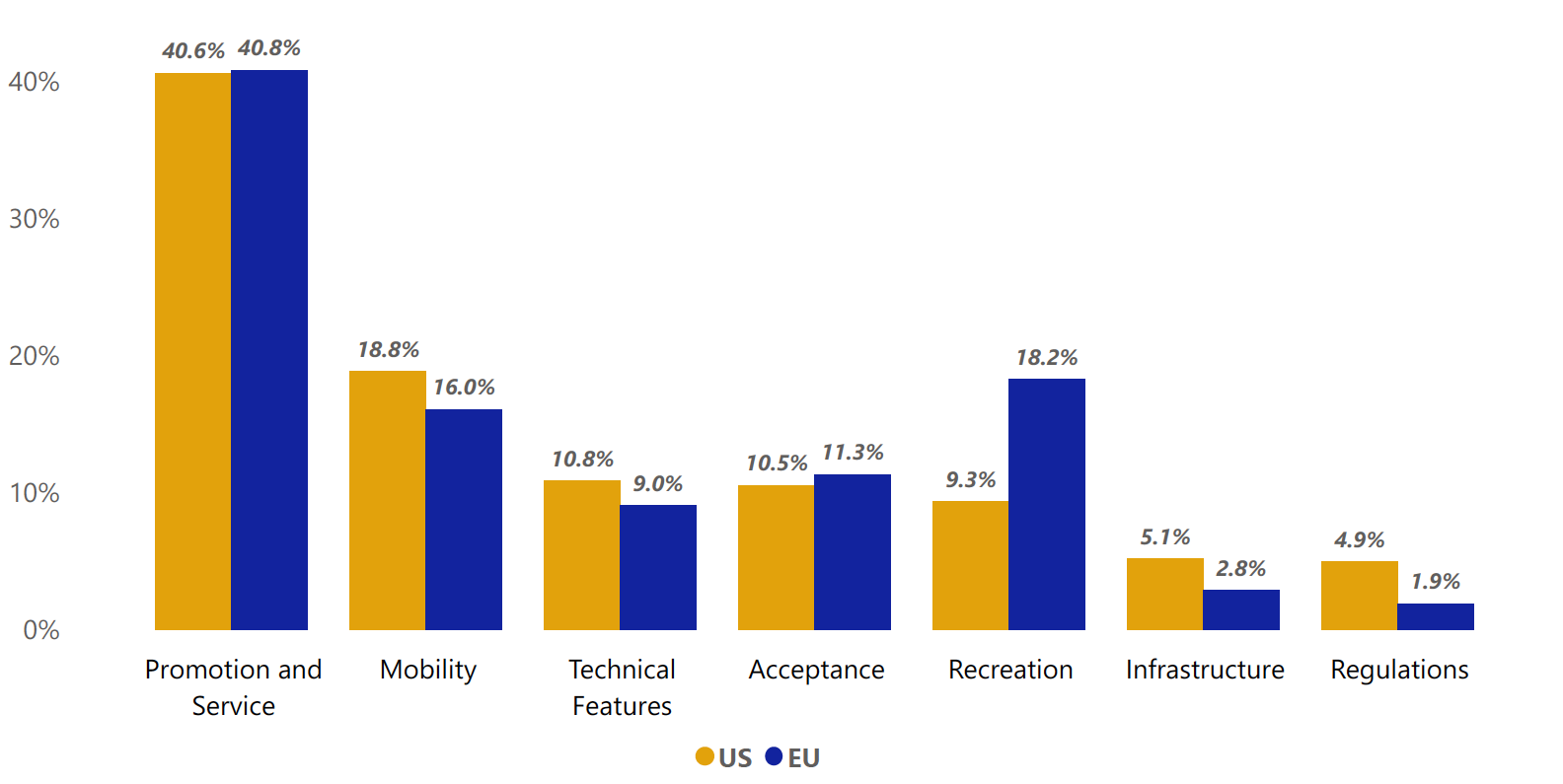}
\caption{Distribution of tweets in the US and EU across seven topics.}
\label{fig:topics}
\end{figure}

Figure \ref{fig:topics} shows the distribution of the topics in the US. The distribution was as follows: promotion and service was the largest topic with 650 tweets (40.6\%), mobility had 301 (18.8\%) tweets, technical features had 173 (10.8\%) tweets, public acceptance had 168 (10.5\%) tweets, recreation had 149 (9.3\%) tweets, infrastructure had 82 (5.1\%) tweets, and the remaining 79 (4.9\%) represented the regulations topic. 

Promotion and service is the largest category we identified, which accounts for 40.6\% in the US. This tends to be consistent with the fact that many of the tweets were generated by micro-mobility companies, who wanted to promote their sales. This topic includes manufacturer updates about new releases, dealerships' advertisements and promotions (e.g., \textit{``Airwheel E3  backpacking  ebike, with a radical design, makes a good start for integration of intelligent vehicles into popular daily  life.''}), as well as tweets related to new customers (e.g., \textit{`` I'm also trying to decide which cargo ebike to buy. I even made a spreadsheet to compare. Haha. Here is a video I created so if you're still looking, you take a look''}). Furthermore, this category also includes  customer service (e.g., \textit{``At least I'll know I can find my next ebike purchase elsewhere. I can't support a company with such a cancellation culture. Bye''}) and financial issues, such as absence of subsidy programs and higher sale prices and rental costs (e.g. \textit{`` Any EV subsidy program that doesn't also have an ebike subsidy is a massive policy failure''}). This topic contains a large number of tweets, indicating that manufacturers and dealers actively use the social media to promote their products and services.

The mobility topic describes the benefits and opportunities of micro-mobility brought, such as the sustainability and efficiency of e-bikes and e-scooters compared to cars (e.g., \textit{``Transportation revolutions  Driverless  EV Shared Mobility together can cut CO2 80 percent'', `` Yes and an ebike is sometimes (most of the time!) the fastest way to get around a city - hello dedicated lane- so it could improve efficiency for the busiest people''}). At the same time, people also shared their experience of using e-bikes and e-scooters as a primary way of transportation (e.g., \textit{``We are a 1 car 2 ebike household. The car sits most days, and kids get dropped of at daycare on e cargobike. A bike doesn't have to replace 100\% of trips to be helpful.'', `` In SF, there are not enough ebikes. I would be very curious about the take rate. I assumed that EVERYONE would choose ebike if there was a choice at the dock. There is so much opportunity to get more people on board micromobility. But little/no outreach beyond…''}). Additionally, this topic discusses the safety benefits (e.g., \textit{``I feel waaay safer going consistently 15-20mph on my ebike than I do doing 9-13mph on a normal one. The better acceleration alone makes city streets feel much safer''}) and concerns (e.g., \textit{``All these 6' wide bike lanes will quickly become too narrow with widespread ebike adoption. 20mph is really fast on a bike in the city.''}) of micro-mobility devices.

The topic of technical features covers the discussions related to micro-mobility devices, such as body (i.e, breaks, weight, cargo space) (e.g., \textit{``Modern hydraulic discs are so great. Also good choice considering ebike speeds.'',``Weight was a main factor in choosing my Trek over an ebike right now. Some days it's hard enough carrying my XXL sized Trek up 3 flights of stairs. Can't imagine doing that with an ebike just yet. But seeing its usefulness on days when I wanna bike run.'',``Nice design but no basket or storage area? Hard to invest in an eBike that I can't do family shopping with.'' }), drive system (i.e, battery, power, speed) (e.g., \textit{``Batteries are also affected by the cold. Straight out of the house, the ebike would zip along at its 28 mph limit. But once it cooled down to freezing it would max out at 23 mph.'',``Most ebikes will struggle with a grade that high (20\%!) And it's hard to know if it would provide any notable boost at all, weak motors may burn. But a mid-drive ebike with a 750-1000watt motor would make it, you'll still have to pedal but it should be notably easier.'',``In my exp the average speed of an ebike is higher than a reg bike bc ebikes accelerate to their top speed more quickly. Riding at top speed for more time means the rider faces a lot more risk if they crash.''}.), and effects of different e-bike classes that were defined by government based on its characteristics in order to regulate the usage  (e.g., ``Weight, speed, but also throttle sensitivity (especially on pedal assist class 1 bike)''). 

Acceptance contains positive and supporting tweets about micro-mobility and expresses public attitudes and preferences towards the micro-mobility devices, especially e-bikes and e-scooters, and its integration into the transportation system (e.g., \textit{``I see the ebike as more a car replacement, I've still got a non electric road bike I plan to still use for run rides with friends at the weekend''}).

Recreation refers to tweets about fun, leisure, tours, and sports. For instance, people discussed and shared stories about their active lifestyles with e-bikes and e-scooters with trips, such as morning cycling and weekend sightseeing tours. Example tweets include \textit{``Beautiful views riding around North Portland today. We rode all the way up and over the St. John's Bridge!'', ``We passed a lady in her 70s on an ebike going through the rollers on 36 Saturday. She had a huge smile on her face and was cruising along. Was awesome'', ``Today's incredible ride from Ullapool harbour involved grit and determination at points!! ?? But Wow, the views more than made up for it!!''}.

For the infrastructure topic, we identified tweets related to road structures and sharing services that supported the usage of micro-mobility devices. These covered the discussions about necessity of new road architectures that included separated lanes for such type of transportation (e.g., \textit{``Oh, and ebike use can't expand to its full potential without infrastructure changes to make non-car travel safer.''}) and necessity of additional pick-up, drop-off, and parking zones (e.g., \textit{``officially released a map of the designated parking spaces for electric scooters! Many thanks to our community for all the support. SMU ? PonyUp ElectricScooter''}), as well as updates about new micro-mobility hubs (e.g., \textit{``Completing 's first of two  MicroMobility Parking Hubs watch for scooters  ebikes this  First Friday''}).\\

The regulations topic discusses the laws, rules, and the traffic restrictions for different type of micro-mobility vehicles. Example tweets are \textit{``economy mode is limited to a top speed of 20 mph, 1,000 watts to comply with California ebike regulations.'',``We go by bicycle/ebike laws for now. There are no real laws regarding the EUC's. Its always good to have a light if riding at night.'', ``I can't park my motorcycle inside my office! USA needs a uniform law limiting top ebike assist speed. EU is 15.5 mph.'',``This passed in 2005 and will be reconsidered along with other regulation over the course of the  dockless  shared mobility pilot'',``BMW and Mercedes limit their cars to 155 mph. Meanwhile, the government limits my ebike to 28 mph so I don't hurt someone.''}

\subsubsection{EU results} Using the labeled tweets from the US as ground truth, we fine-tuned six pre-trained models from transformers, and tested with different parameters. Table \ref{table:perf} summarizes the f1-scores for all the tested models for predicting topics, where we used 5-fold cross-validation with max sequence length $= 256$, learning rate $= 3\times 10^{-5}$ and epoch $= 10$. The Twitter-roBERTa model had the best performance among all the models resulting in f1-score $= .90$ prediction performance. Then, we used the trained Twitter-roBERTa model to predict the topics of the micro-mobility tweets posted in the EU (see Figure \ref{fig:process}).

The EU results were generally similar to the US, but with some distinct patterns. First, both promotion and service was the largest topic that social media was widely used for promotion and service updates in EU also with 40.8 \% of the tweets. Respectively, mobility accounted for 16.0\%, acceptance accounted for 11.3\%, technical features accounted for 9.0\%, infrastructure accounted for 2.8\%, and regulations accounted for 1.9\% (see Figure \ref{fig:topics}).
In the EU, the distribution of tweets for mobility, technical features and acceptance topics was slightly different from the US result. Therefore, we proceeded with a detailed analysis for each topic independently and compared the findings in the discussion section.
However, the percentage of recreation related tweets was 18.2\%, which was almost twice as much as that in the US, indicating that micro-mobility usage for recreation purposes was more developed in the EU than that in the US. 
Also, infrastructure and regulations in the US were much more in the US than in the EU. In general, EU has clearly established laws for micro-mobility devices while the US is still in the process of defining laws and rules regarding the usage of micro-mobility devices and their subcategories on roads or in public areas.

\subsection{Sentiment analysis}

\subsubsection{US results} With regard to sentiment analysis, results indicated that 47.4\% of all the US tweets were assessed as neutral,  43.4\%  were positive and 9.0\% were negative (see Figure \ref{fig:sent}), indicating that almost half of the users tended to favor the current e-bikes and e-scooters.

\subsubsection{EU results} 
\begin{figure}[t]
\centering
\includegraphics[width=.7\linewidth]{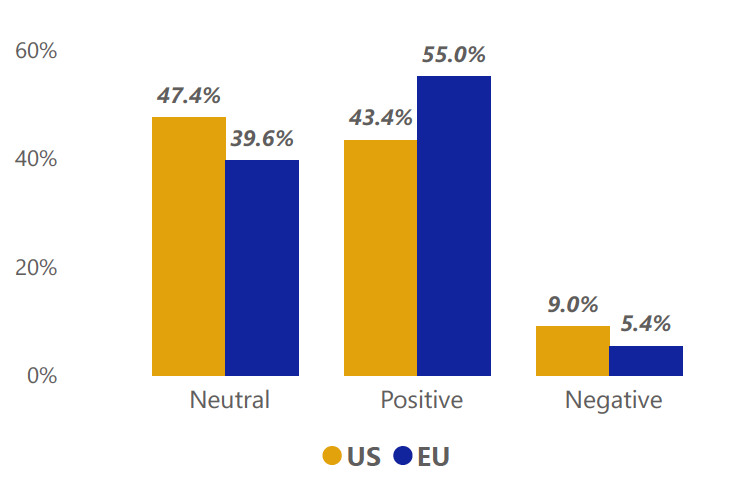}
\caption{Distribution of positive, neutral and negative tweets in the US and EU.}
\label{fig:sent}
\end{figure}

We replicated the supervised learning and prediction process to predict sentiment polarities in the EU using sentiment labels in the US data. Here, we replaced Twitter-roBERTa base model with the sentiment-oriented version since it was specifically pre-trained for sentiment evaluation. Table \ref{table:perf} illustrates the prediction performance for models and shows that Twitter-roBERTa model outperforms others by having f1-score $= 0.91$ in prediction.
In the EU, prediction results showed that majority of tweets were categorized as positive and was accounted for 55.0\% of EU tweets. Since in the US lots of tweets were neutral (47.4\%), people in the EU had more positive experience using micro-mobility devices, and were more satisfied with micro-mobility integration than in the US. Respectively, 39.6\% were predicted as neutral, and 5.4\% as negative (see Figure \ref{fig:sent}) indicating that in the EU people were facing problems 60\% less than in the US.

\begin{table}[!ht]
 \caption{The f1-scores and ROC\_AUC scores of evaluated models for the topic and sentiment }
 \centering
 \begin{adjustbox}{width=1\textwidth}
\begin{tabular}{lllllllllllll}
\hline
\hline
          & \multicolumn{2}{c}{BERT} & \multicolumn{2}{c}{BERTweet} & \multicolumn{2}{c}{DistilBert} & \multicolumn{2}{c}{DistilRoberta} & \multicolumn{2}{c}{Twitter-Roberta} & \multicolumn{2}{c}{XLNet} \\
          \hline
          & f1-score    & roc\_auc   & f1-score      & roc\_auc     & f1-score       & roc\_auc      & f1-score        & roc\_auc        & f1-score         & roc\_auc         & f1-score    & roc\_auc    \\
          \hline
Topic     & 0.79        & 0.91       & 0.88          & 0.89         & 0.88           & 0.87          & 0.84            & 0.90            & 0.90             & 0.93
& 0.82        & 0.83        \\
\hline
Sentiment & 0.89        & 0.73       & 0.84          & 0.93         & 0.86           & 0.94          & 0.85            & 0.93            & 0.91             & 0.98             & 0.80        & 0.90       \\
\hline

\end{tabular}
\end{adjustbox}
\label{table:perf}
\end{table}



\begin{figure}
\centering
\includegraphics[width=1\linewidth]{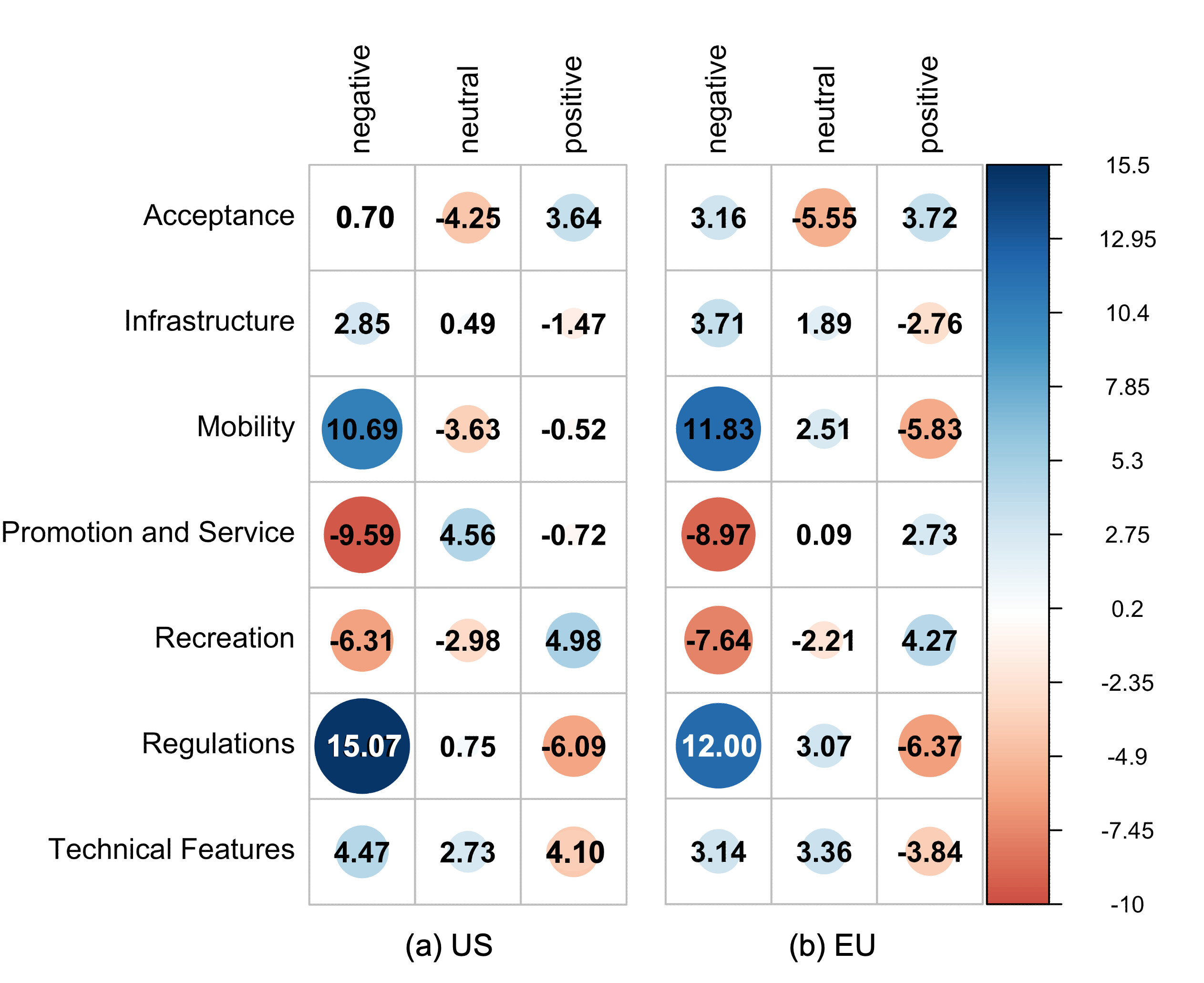}
\caption{Associations between topics and sentiments.}
\label{fig:chi}
\end{figure}

In addition, we conducted a Chi-Square analysis to determine the associations between sentiment polarities and topics. Although we could not find statistical difference per topic (${\chi}^2 = 1.16, p = .978$) and per sentiment (${\chi}^2 = 3.04, p = .218$ between the US and EU, there was a significant difference within each region.

In the US, the results showed that topics were significantly associated with their sentiments (${\chi}^2 = 665.90, p < .001$).  The post-hoc analysis showed that mobility had significantly more ($p < .001$) negative tweets, and recreation had significantly fewer negative tweets (compared with Infrastructure ($p = .011$) and the remaining topics ($p < .001$)). The strong association between promotion and service with neutral sentiment in Figure \ref{fig:chi} indicated that the topic included the largest amount of neutral tweets than other topics ($p < .000$). Regarding to the positive sentiment, distribution of tweets was significantly different among all topics ($p < .013$). While recreation ($p < .000$) and promotion and service ($p < .000$) were leading by a significantly larger amount of positive tweets, the association of regulations, infrastructure and technical features with positive sentiment was significantly weak ($p < .000$).

We also found significant differences for associations between topics and sentiments in the EU (${\chi}^2 = 659.19, p < .001$). Post-hoc analysis showed that the positive tweets were strongly associated with recreation and acceptance than with other topics. The neutral sentiment was mainly associated with technical features and regulations. Nevertheless, mobility included the largest amount of negative tweets than other, the negative sentiment had the strongest association with regulations topic followed by infrastructure.
Figure \ref{fig:chi} represents the topic and sentiment associations in each region based on Pearson residuals shown in circles. The positive values, demonstrated in blue, indicated positive association between topic and sentiment, and the larger the value was, the stronger the association was. For instance, regulations were strongly associated with negative sentiment and were weakly associated with the neutral sentiment. Likewise, the negative values, demonstrated in red, indicate the negative associations and the size showed the strength of association. For instance, mobility was negatively associated with the neutral sentiment in the EU.

\section{DISCUSSIONS}
With the BertTopic model and manual labelling, we identified seven topics that people from the US and the EU mainly discussed in the social media. We also found that Twitter-Roberta and Twitter-Roberta (sentiment) models performed the best on our data. The sections below provide detailed explanations of results for each region and their similarities and differences.

\subsection{US}

Our proposed method identified relevant topics that positively impacted micro-mobility integration with the transportation system. Results showed that mobility was the first major topic acknowledged in the US tweets as a major benefit of micro-mobility. In particular, people highlighted the cost-efficiency of using micro-mobility devices compared to cars and other transportation modes (e.g., \textit{``Oddly enough.. I drive my car about 75\% less since I got my ebike. Between that and ride-sharing, not sure you need to own a car anymore. That's in CA, at least. Ebikes might take down the entire auto industry''}). As a part of mobility, the sharing option also had an important role for long-distance rides because users did not need to worry about charging. The literature review also supported this finding as recent years people switched more from public transportation to micro-mobility devices and shared mobility modes, particularly during and COVID-19 pandemic \citep{nikitas2021cycling}. 

The second topic with a major positive sentiment was technical features. The design concept, such as folding frames, was appreciated by many people as it was easy to carry and store. Also, the pedal assistance feature was valued since users could take over control of the power used by the e-bike's drive system and save battery more mileages. (e.g.,\emph{``I can go on a slow roll group ride on an ebike with 100 other regular bikes and nobody even notices it's electric. Pretty sure I could not do that with a motorcycle''}). 

Third, in the acceptance topic, we found that people tended to use e-bikes to maintain an active and healthy lifestyle with less efforts compared to regular bikes  (e.g., \textit{``I think eBikes could be huge in the US, because we have a) hills and b) humidity. Specifically, I suspect many professionals can't (or don't want to) shower at work to rinse off after biking to the office, while an eBike would allow them to commute without getting sweaty''}). In addition, recent design updates satisfied critical requirements to replace cars. For example, e-bikes with cargo racks could easily haul other passengers and groceries, which were previously mentioned as a main concern of replacing cars with micro-mobility devices.

Fourth, using e-bikes and e-scooters for recreational purposes became popular among people in the US. E-bike riders were more motivated to spend leisure time having short trips in nearby areas or go for weekend trips in groups while regular bicycles were used for utilitarian purposes (e.g., \emph{``Another Friday, another full and fun weekend of valuable  eTrike time ahead of us!! Happy Friday everyone!''}).


However, in some topics, people often faced challenges by using micro-mobility vehicles in the US. In detail, with regard to technical features, the battery issues were mostly discussed in social media pointing out that cold weather significantly influenced battery lifetime (e.g., \textit{``Batteries are also affected by the cold. Straight out of the house, the ebike would zip along at its 28 mph limit. But once it cooled down to freezing it would max out at 23 mph.''}), causing faster discharge or complete damage. This resulted in extra expenses which disappointed users. Previous studies  \citep{macarthur2018north, loustric2020} reported concerns about distance per battery, but there was no indication about battery issues under different weather conditions.
The heavyweight was the second major factor mentioned in the tweets which was consistent with the literature. Though it is well known that e-bikes are heavier than regular bicycles due to the motor system and battery, riders were expecting lighter e-bikes in the market (e.g., \emph{``Weight was a main factor in choosing my Trek over an ebike right now. Some days it's hard enough carrying my XXL sized Trek up 3 flights of stairs...''}). 

Second, the financial burden as mentioned in the promotion and service category was another concern. The e-bike purchase price as well as the rent and maintenance cost were notably higher than a regular bicycle and were not affordable for many middle-income users (e.g., \textit{``\$15/mo + \$2 per ebike ride I'm disappointed by the rollout of ebike pricing in NYC. I know we're going to need to pay more for ebikes but the communication and structure of this change misses the mark.''}). The major complaint was that unlike the electric vehicles, there were no subsidy programs for e-bikes or e-scooters offered by the government (e.g., \textit{``on to the tax code. If you buy a luxury EV the government will cut you a check for \$7k. Almost no where has subsidies for folks who want to buy an ebike''}). Although \citet{ling2017differences}, already pointed out the financial issues (i.e., price and maintenance cost) as a major barrier, our findings provided more details about people's expectations.

Third, infrastructure-related problems hamper the use and integration of micro-mobility devices. The main problems were a lack of dedicated lanes (mentioned by both car drivers and micro-mobility device owners), safe parking zones, and charging stations (e.g., \emph{``We should absolutely convert a car lane. Replacing car infra with ebike infra is exactly the climate action we need'', ``Addressing shared mobility when planning for the future of transportation is important. When planning for transportation are we adjusting requirements for parking?''}). \citet{abduljabbar2021role} also pointed out the parking safety concerns and mentioned that availability of micro-mobility devices (i.e., e-bikes and e-scooters) was a problem for micro-mobility development. Regulation issues were also highlighted as a barrier. We found contradicted opinions about speed limit regulations, as for owners, it limited the mobility advantage they were getting from micro-mobility devices while pedestrians were complaining that the defined limits were not safe enough to prevent them from severe injuries (e.g., \textit{``So I can now legally ride my cargo carrying ebike that weighs less than 80lbs up to 12 mph on the sidewalk? sweet. Also, how much is the value of a dead/injured pedestrian or cyclist? \$1000? \$1500? \$2000?''}). Moreover, there were no federal laws defined to regulate accidents that included e-bikes or e-scooters. In contrast, previous studies only \citep{aman2021listen, boglietti2021survey} included concerns related to the lack of speed legislation and regulations with other road users. 

\subsection{EU}
In the EU, micro-mobility devices, particularly e-bikes, were widely accepted among the elderly and people in special medical conditions. In the acceptance topic, we found that people with various health problems (e.g., \textit{``Sometimes walking is harder for people than cycling like my wife who had spinal surgery, her ebike allows her to get around easier''}) switched to e-bikes by doctors' suggestion and found it extremely helpful. Meantime, people also paid more attention to the fact that these devices were an eco-friendly transportation mode (e.g., \textit{``We won't stop working towards our goal to make our everyday life more \#sustainable and our world a more livable place. Thanks to our electric micromobility services we will reduce the carbon emissions caused by existing urban transportation networks and make them more efficient.''}). 
Unlike the US, mobility-related tweets showed that time-efficiency was a key factor in using micro-mobility devices for utilitarian purposes (e.g., \textit{``Compared to using our PHEV (as an electric car) then using the ebike is saving ~3.5kWh for the return trip to the office or approx 1kgCO2 and of course better for me and taking up less road space etc. So assuming I manage 100 return trips to work per year then that 100kgCO2...!''}). For e-bikes, people also found it safer to use in the winter and in the long distance rides (i.e., more than 20 miles) rather than regular bicycles.
The EU had a well-developed recreation network. Activities, such as multi-city e-bike touring, were common among different countries (e.g., \textit{``We're excited to announce as well as \#ebike hire, we now do guided electricbike tours. 
We have lots of tours in the pipeline across the South. If you have any special requests, we'd love to hear from you.''}), and because many cafes and restaurants were providing charging points, it was convenient to use for long-distance recreation purposes (e.g., \textit{``Interesting concept.  batteries being charged in a local bakery. Can I please get some background (prices, terms and conditions) please? \#smartcity \#micromobility \#escooter''}).

However, the EU needs improvements for the expansion of micro-mobility. Although the infrastructure supports micro-mobility usage in many countries, there was still a need to redesign the roads for safer integration (e.g., \textit{``Are you any more likely to be hit by an ebike than a regular bike in the UK? I imagine same policy applies. No number plates required here. I suppose the separated, dedicated lanes here mean collisions rarely occur.''}). Moreover, there was a lack of maintenance services for damaged stations. Another major issue was about parking stations' security, where people mentioned that their parked e-bikes were stolen (e.g., \textit{``This seemed sensible until I remembered that an eBike is so easily stolen. I'd love to go to the shops on a bike return with wares in a rucksack. But there's no way at all I can leave an eBike for more than a few seconds even while locked. Secure storage is an eBike necessity.''}). There is a lack of such information in the literature.
Users were also facing technical issues, such as battery overheating that resulted in shorter lifetime than manufacturers mentioned (e.g., \textit{``Don't forget your \#ebike \#batteries in this weather. Remember the cold will also affect the range you get out of your battery, so make sure you keep them topped up too.''}). Furthermore, some complaints were about maximum speed capability that weight (i.e., obese people or carrying extra batteries) notably reduced the speed (e.g., \textit{``My ebike slows me down, as it is slower than my other bikes due to the weight of the motor and battery.''}). Finally, issues related to regulating e-bikes and e-scooters with unclear definitions and legalization were not addressed, resulting in penalties (e.g., \textit{``Electric scooters are illegal in the UK and users face a £300 fixed penalty and 6 points on their driving license! I am surprised as I see people using these daily!.''}). 

\subsection{Design a Sustainable Micro-mobility Future}
The trends and challenges of micro-mobility across the US and the EU help better understand different priorities in designing a sustainable micro-mobility future in several aspects. 
First, micro-mobility vehicles reduce people's dependence on traditional vehicles (especially for short distance travels) by as much as 30\% \citep{fitt2019scooter}, which is more evident in the EU and now many big cities in the US. This will improve environmental sustainability by reducing carbon dioxide emissions. Further work needs to be done in this aspect in both areas, such as designing and building more related transportation infrastructure across different cities, including dedicated lanes, safe parking zones, charging stations, and repair shops. Related laws and regulations about micro-mobility should also be proposed and defined. 
Second, micro-mobility and its sharing programs should be well designed to connect the public transportation system to solve the first and last mile problems so that users can travel across a longer distance with flexibility and low costs. Such design might well address systemic issues around transportation equality. The financial burden was found as one concern compared to traditional bicycles. In this aspect, government subsidy programs like those for electric cars should be considered for micro-mobility devices as well. Another perspective is through innovative design, such as more lightweight materials, more efficient battery technologies, autonomous damage sensors, and puncture proof tires for more sustainable rides.
Third, as shown in the EU, many elderly people with different health conditions were still able to maintain an active and healthy lifestyle through micro-mobility. Through proper promotion by commercials and recommendations by doctors, not only younger people, but also the elderly are able to make everyday life more sustainable. 

\subsection{Limitations and Future Work}
This study also has limitations that can be examined in future studies.
First, the data was collected only from the Twitter social media platform since other possible sources could not provide geographic information about data. The vast majority of the collected data did not include geographic information and were excluded from analysis. Since the study was aimed to compare the trends and challenges in the EU and the US, all the tweets from unknown locations were excluded from the analysis. Thus, it might have selection biases in reporting the results based only with the current collected Twitter data. Therefore, for future research,  1) additional sources, such as articles from big media agencies or comments from YouTube videos can be collected, 2) additional data inspection process is needed in order to identify the possible origin before proceeding to the analysis, and 3) further investigations are needed to understand the possible issues especially in the EU and its effects on the overall trend.

Second, we filtered and analyzed tweets in English which remarkably reduced the amount of EU tweets. One possible reason could be that English is not a national language for most of the EU countries. Therefore, it is worthwhile to include tweets in countries' native languages for further investigation.
Third, only US tweets, which had imbalanced topics, were used for the supervised models' training which might result in biased models and eventually affect the prediction accuracy. The labels of the models were also provided by subjective labeling. In the future, more data with relatively balanced labels and more reliable human raters should be used to train the model to improve the overall performance in the future. 

\section{CONCLUSION}

In this study, we investigated the micro-mobility trends and challenges in the US and the EU and aimed to identify the differences between these two regions using tweet data. We applied the state-of-the-art NLP methods to analyse the collected data and extract main areas of people's interests in micro-mobility, and explored the polarity of their opinions related to particular topics. We identified seven major topics (i.e., promotion and service, mobility, technical features, acceptance, recreation, infrastructure and regulations) based on Twitter data set that contained micro-mobility information. Next, we explored the sentiment polarity (i.e., positive, neutral, and negative) of the data and determined their distributions across the two regions. We found that the EU was dominated by positive opinions about micro-mobility. However, topics, such as infrastructure, regulations and technical features were dominated by negative opinions in both regions. These findings showed that our method complemented the challenges and trends that were found in the literature, and provided better understanding from the users' point of view expressed in social media.

\subsection{ACKNOWLEDGMENT}
This work was supported by the University of Michigan Mcity. The views expressed are those of the authors and do not reflect the official policy or position of Honda.
\bibliography{HFES-bibliography}

\begin{thebibliography}{}

\bibitem [\protect \citeauthoryear {%
Abduljabbar%
, Liyanage%
\BCBL {}\ \BBA {} Dia%
}{%
Abduljabbar%
\ \protect \BOthers {.}}{%
{\protect \APACyear {2021}}%
}]{%
abduljabbar2021role}
\APACinsertmetastar {%
abduljabbar2021role}%
\begin{APACrefauthors}%
Abduljabbar, R\BPBI L.%
, Liyanage, S.%
\BCBL {}\ \BBA {} Dia, H.%
\end{APACrefauthors}%
\unskip\
\newblock
\APACrefYearMonthDay{2021}{}{}.
\newblock
{\BBOQ}\APACrefatitle {The role of micro-mobility in shaping sustainable
  cities: A systematic literature review} {The role of micro-mobility in
  shaping sustainable cities: A systematic literature review}.{\BBCQ}
\newblock
\APACjournalVolNumPages{Transportation research part D: transport and
  environment}{92}{}{102734}.
\PrintBackRefs{\CurrentBib}

\bibitem [\protect \citeauthoryear {%
Aman%
, Smith-Colin%
\BCBL {}\ \BBA {} Zhang%
}{%
Aman%
\ \protect \BOthers {.}}{%
{\protect \APACyear {2021}}%
}]{%
aman2021listen}
\APACinsertmetastar {%
aman2021listen}%
\begin{APACrefauthors}%
Aman, J\BPBI J.%
, Smith-Colin, J.%
\BCBL {}\ \BBA {} Zhang, W.%
\end{APACrefauthors}%
\unskip\
\newblock
\APACrefYearMonthDay{2021}{}{}.
\newblock
{\BBOQ}\APACrefatitle {Listen to E-scooter riders: Mining rider satisfaction
  factors from app store reviews} {Listen to e-scooter riders: Mining rider
  satisfaction factors from app store reviews}.{\BBCQ}
\newblock
\APACjournalVolNumPages{Transportation research part D: transport and
  environment}{95}{}{102856}.
\PrintBackRefs{\CurrentBib}

\bibitem [\protect \citeauthoryear {%
Ayoub%
\ \protect \BOthers {.}}{%
Ayoub%
\ \protect \BOthers {.}}{%
{\protect \APACyear {2022}}%
}]{%
ayoub2022cause}
\APACinsertmetastar {%
ayoub2022cause}%
\begin{APACrefauthors}%
Ayoub, J.%
, Wang, Z.%
, Li, M.%
, Guo, H.%
, Sherony, R.%
, Bao, S.%
\BCBL {}\ \BBA {} Zhou, F.%
\end{APACrefauthors}%
\unskip\
\newblock
\APACrefYearMonthDay{2022}{}{}.
\newblock
{\BBOQ}\APACrefatitle {Cause-and-Effect Analysis of ADAS: A Comparison Study
  between Literature Review and Complaint Data} {Cause-and-effect analysis of
  adas: A comparison study between literature review and complaint
  data}.{\BBCQ}
\newblock
\BIn{} \APACrefbtitle {Proceedings of the 14th International Conference on
  Automotive User Interfaces and Interactive Vehicular Applications}
  {Proceedings of the 14th international conference on automotive user
  interfaces and interactive vehicular applications}\ (\BPGS\ 139--149).
\PrintBackRefs{\CurrentBib}

\bibitem [\protect \citeauthoryear {%
Barbieri%
, Camacho-Collados%
, Neves%
\BCBL {}\ \BBA {} Espinosa-Anke%
}{%
Barbieri%
\ \protect \BOthers {.}}{%
{\protect \APACyear {2020}}%
}]{%
barbieri2020tweeteval}
\APACinsertmetastar {%
barbieri2020tweeteval}%
\begin{APACrefauthors}%
Barbieri, F.%
, Camacho-Collados, J.%
, Neves, L.%
\BCBL {}\ \BBA {} Espinosa-Anke, L.%
\end{APACrefauthors}%
\unskip\
\newblock
\APACrefYearMonthDay{2020}{}{}.
\newblock
{\BBOQ}\APACrefatitle {Tweeteval: Unified benchmark and comparative evaluation
  for tweet classification} {Tweeteval: Unified benchmark and comparative
  evaluation for tweet classification}.{\BBCQ}
\newblock
\APACjournalVolNumPages{arXiv preprint arXiv:2010.12421}{}{}{}.
\PrintBackRefs{\CurrentBib}

\bibitem [\protect \citeauthoryear {%
Blei%
, Ng%
\BCBL {}\ \BBA {} Jordan%
}{%
Blei%
\ \protect \BOthers {.}}{%
{\protect \APACyear {2003}}%
}]{%
blei2003LDA}
\APACinsertmetastar {%
blei2003LDA}%
\begin{APACrefauthors}%
Blei, D\BPBI M.%
, Ng, A\BPBI Y.%
\BCBL {}\ \BBA {} Jordan, M\BPBI I.%
\end{APACrefauthors}%
\unskip\
\newblock
\APACrefYearMonthDay{2003}{mar}{}.
\newblock
{\BBOQ}\APACrefatitle {Latent Dirichlet Allocation} {Latent dirichlet
  allocation}.{\BBCQ}
\newblock
\APACjournalVolNumPages{J. Mach. Learn. Res.}{3}{}{993–1022}.
\PrintBackRefs{\CurrentBib}

\bibitem [\protect \citeauthoryear {%
Blomberg%
, Rosenkrantz%
, Lippert%
\BCBL {}\ \BBA {} Christensen%
}{%
Blomberg%
\ \protect \BOthers {.}}{%
{\protect \APACyear {2019}}%
}]{%
blomberg2019injury}
\APACinsertmetastar {%
blomberg2019injury}%
\begin{APACrefauthors}%
Blomberg, S\BPBI N\BPBI F.%
, Rosenkrantz, O\BPBI C\BPBI M.%
, Lippert, F.%
\BCBL {}\ \BBA {} Christensen, H\BPBI C.%
\end{APACrefauthors}%
\unskip\
\newblock
\APACrefYearMonthDay{2019}{}{}.
\newblock
{\BBOQ}\APACrefatitle {Injury from electric scooters in Copenhagen: a
  retrospective cohort study} {Injury from electric scooters in copenhagen: a
  retrospective cohort study}.{\BBCQ}
\newblock
\APACjournalVolNumPages{BMJ open}{9}{12}{e033988}.
\PrintBackRefs{\CurrentBib}

\bibitem [\protect \citeauthoryear {%
Boglietti%
, Barabino%
\BCBL {}\ \BBA {} Maternini%
}{%
Boglietti%
\ \protect \BOthers {.}}{%
{\protect \APACyear {2021}}%
}]{%
boglietti2021survey}
\APACinsertmetastar {%
boglietti2021survey}%
\begin{APACrefauthors}%
Boglietti, S.%
, Barabino, B.%
\BCBL {}\ \BBA {} Maternini, G.%
\end{APACrefauthors}%
\unskip\
\newblock
\APACrefYearMonthDay{2021}{}{}.
\newblock
{\BBOQ}\APACrefatitle {Survey on e-Powered Micro Personal Mobility Vehicles:
  Exploring Current Issues towards Future Developments} {Survey on e-powered
  micro personal mobility vehicles: Exploring current issues towards future
  developments}.{\BBCQ}
\newblock
\APACjournalVolNumPages{Sustainability}{13}{7}{}.
\newblock
\begin{APACrefDOI} \doi{10.3390/su13073692} \end{APACrefDOI}
\PrintBackRefs{\CurrentBib}

\bibitem [\protect \citeauthoryear {%
Campisi%
, Skoufas%
, Kaltsidis%
\BCBL {}\ \BBA {} Basbas%
}{%
Campisi%
\ \protect \BOthers {.}}{%
{\protect \APACyear {2021}}%
}]{%
campisi2021gender}
\APACinsertmetastar {%
campisi2021gender}%
\begin{APACrefauthors}%
Campisi, T.%
, Skoufas, A.%
, Kaltsidis, A.%
\BCBL {}\ \BBA {} Basbas, S.%
\end{APACrefauthors}%
\unskip\
\newblock
\APACrefYearMonthDay{2021}{}{}.
\newblock
{\BBOQ}\APACrefatitle {Gender equality and E-scooters: Mind the gap! A
  statistical analysis of the Sicily Region, Italy} {Gender equality and
  e-scooters: Mind the gap! a statistical analysis of the sicily region,
  italy}.{\BBCQ}
\newblock
\APACjournalVolNumPages{Social Sciences}{10}{10}{403}.
\PrintBackRefs{\CurrentBib}

\bibitem [\protect \citeauthoryear {%
Dahl%
\ \BBA {} Haug%
}{%
Dahl%
\ \BBA {} Haug%
}{%
{\protect \APACyear {2020}}%
}]{%
dahl2020scooter}
\APACinsertmetastar {%
dahl2020scooter}%
\begin{APACrefauthors}%
Dahl, A.%
\BCBT {}\ \BBA {} Haug, J\BPBI A.%
\end{APACrefauthors}%
\unskip\
\newblock
\APACrefYear{2020}.
\unskip\
\newblock
\APACrefbtitle {E-scooter--convenience versus environment: a green innovation
  value chain analysis of transportation in urban areas}
  {E-scooter--convenience versus environment: a green innovation value chain
  analysis of transportation in urban areas}\ \APACtypeAddressSchool
  {\BUMTh}{}{}.
\unskip\
\newblock
\APACaddressSchool {}{Handelsh{\o}yskolen BI}.
\PrintBackRefs{\CurrentBib}

\bibitem [\protect \citeauthoryear {%
Devlin%
, Chang%
, Lee%
\BCBL {}\ \BBA {} Toutanova%
}{%
Devlin%
\ \protect \BOthers {.}}{%
{\protect \APACyear {2019}}%
}]{%
devlin2019BERT}
\APACinsertmetastar {%
devlin2019BERT}%
\begin{APACrefauthors}%
Devlin, J.%
, Chang, M\BHBI W.%
, Lee, K.%
\BCBL {}\ \BBA {} Toutanova, K.%
\end{APACrefauthors}%
\unskip\
\newblock
\APACrefYearMonthDay{2019}{}{}.
\newblock
{\BBOQ}\APACrefatitle {BERT: Pre-training of Deep Bidirectional Transformers
  for Language Understanding} {Bert: Pre-training of deep bidirectional
  transformers for language understanding}.{\BBCQ}
\newblock
\APACjournalVolNumPages{ArXiv}{abs/1810.04805}{}{}.
\PrintBackRefs{\CurrentBib}

\bibitem [\protect \citeauthoryear {%
Drus%
\ \BBA {} Khalid%
}{%
Drus%
\ \BBA {} Khalid%
}{%
{\protect \APACyear {2019}}%
}]{%
drus2019soc_med}
\APACinsertmetastar {%
drus2019soc_med}%
\begin{APACrefauthors}%
Drus, Z.%
\BCBT {}\ \BBA {} Khalid, H.%
\end{APACrefauthors}%
\unskip\
\newblock
\APACrefYearMonthDay{2019}{}{}.
\newblock
{\BBOQ}\APACrefatitle {Sentiment analysis in social media and its application:
  Systematic literature review} {Sentiment analysis in social media and its
  application: Systematic literature review}.{\BBCQ}
\newblock
\APACjournalVolNumPages{Procedia Computer Science}{161}{}{707--714}.
\PrintBackRefs{\CurrentBib}

\bibitem [\protect \citeauthoryear {%
Fitt%
\ \BBA {} Curl%
}{%
Fitt%
\ \BBA {} Curl%
}{%
{\protect \APACyear {2019}}%
}]{%
fitt2019scooter}
\APACinsertmetastar {%
fitt2019scooter}%
\begin{APACrefauthors}%
Fitt, H.%
\BCBT {}\ \BBA {} Curl, A.%
\end{APACrefauthors}%
\unskip\
\newblock
\APACrefYearMonthDay{2019}{}{}.
\newblock
{\BBOQ}\APACrefatitle {E-scooter use in New Zealand: Insights around some
  frequently asked questions} {E-scooter use in new zealand: Insights around
  some frequently asked questions}.{\BBCQ}
\newblock
\APACjournalVolNumPages{University of Canterbury: Christchurch, New
  Zealand}{}{}{}.
\PrintBackRefs{\CurrentBib}

\bibitem [\protect \citeauthoryear {%
G{\"o}ssling%
}{%
G{\"o}ssling%
}{%
{\protect \APACyear {2020}}%
}]{%
gossling2020integrating}
\APACinsertmetastar {%
gossling2020integrating}%
\begin{APACrefauthors}%
G{\"o}ssling, S.%
\end{APACrefauthors}%
\unskip\
\newblock
\APACrefYearMonthDay{2020}{}{}.
\newblock
{\BBOQ}\APACrefatitle {Integrating e-scooters in urban transportation:
  Problems, policies, and the prospect of system change} {Integrating
  e-scooters in urban transportation: Problems, policies, and the prospect of
  system change}.{\BBCQ}
\newblock
\APACjournalVolNumPages{Transportation Research Part D: Transport and
  Environment}{79}{}{102230}.
\PrintBackRefs{\CurrentBib}

\bibitem [\protect \citeauthoryear {%
Grootendorst%
}{%
Grootendorst%
}{%
{\protect \APACyear {2020}}%
}]{%
grootendorst2020bertopic}
\APACinsertmetastar {%
grootendorst2020bertopic}%
\begin{APACrefauthors}%
Grootendorst, M.%
\end{APACrefauthors}%
\unskip\
\newblock
\APACrefYearMonthDay{2020}{}{}.
\newblock
\APACrefbtitle {BERTopic: Leveraging BERT and c-TF-IDF to create easily
  interpretable topics.} {Bertopic: Leveraging bert and c-tf-idf to create
  easily interpretable topics.}
\newblock
\APACaddressPublisher{}{Zenodo}.
\newblock
\begin{APACrefURL} \url{https://doi.org/10.5281/zenodo.4381785}
  \end{APACrefURL}
\newblock
\begin{APACrefDOI} \doi{10.5281/zenodo.4381785} \end{APACrefDOI}
\PrintBackRefs{\CurrentBib}

\bibitem [\protect \citeauthoryear {%
Handy%
\ \BBA {} Fitch%
}{%
Handy%
\ \BBA {} Fitch%
}{%
{\protect \APACyear {2022}}%
}]{%
handy2022}
\APACinsertmetastar {%
handy2022}%
\begin{APACrefauthors}%
Handy, S\BPBI L.%
\BCBT {}\ \BBA {} Fitch, D\BPBI T.%
\end{APACrefauthors}%
\unskip\
\newblock
\APACrefYearMonthDay{2022}{}{}.
\newblock
{\BBOQ}\APACrefatitle {Can an e-bike share system increase awareness and
  consideration of e-bikes as a commute mode? Results from a natural
  experiment} {Can an e-bike share system increase awareness and consideration
  of e-bikes as a commute mode? results from a natural experiment}.{\BBCQ}
\newblock
\APACjournalVolNumPages{International Journal of Sustainable
  Transportation}{16}{1}{34--44}.
\PrintBackRefs{\CurrentBib}

\bibitem [\protect \citeauthoryear {%
Haustein%
\ \BBA {} M{\o}ller%
}{%
Haustein%
\ \BBA {} M{\o}ller%
}{%
{\protect \APACyear {2016}}%
}]{%
haustein2016bike}
\APACinsertmetastar {%
haustein2016bike}%
\begin{APACrefauthors}%
Haustein, S.%
\BCBT {}\ \BBA {} M{\o}ller, M.%
\end{APACrefauthors}%
\unskip\
\newblock
\APACrefYearMonthDay{2016}{}{}.
\newblock
{\BBOQ}\APACrefatitle {E-bike safety: individual-level factors and incident
  characteristics} {E-bike safety: individual-level factors and incident
  characteristics}.{\BBCQ}
\newblock
\APACjournalVolNumPages{Journal of Transport \& Health}{3}{3}{386--394}.
\PrintBackRefs{\CurrentBib}

\bibitem [\protect \citeauthoryear {%
Hutto%
\ \BBA {} Gilbert%
}{%
Hutto%
\ \BBA {} Gilbert%
}{%
{\protect \APACyear {2014}}%
}]{%
hutto2014vader}
\APACinsertmetastar {%
hutto2014vader}%
\begin{APACrefauthors}%
Hutto, C.%
\BCBT {}\ \BBA {} Gilbert, E.%
\end{APACrefauthors}%
\unskip\
\newblock
\APACrefYearMonthDay{2014}{}{}.
\newblock
{\BBOQ}\APACrefatitle {Vader: A parsimonious rule-based model for sentiment
  analysis of social media text} {Vader: A parsimonious rule-based model for
  sentiment analysis of social media text}.{\BBCQ}
\newblock
\BIn{} \APACrefbtitle {Proceedings of the international AAAI conference on web
  and social media} {Proceedings of the international aaai conference on web
  and social media}\ (\BVOL~8, \BPGS\ 216--225).
\PrintBackRefs{\CurrentBib}

\bibitem [\protect \citeauthoryear {%
Jiao%
, Luo%
, Malmqvist%
, Johan%
\BCBL {}\ \BBA {} Summers%
}{%
Jiao%
\ \protect \BOthers {.}}{%
{\protect \APACyear {2022}}%
}]{%
jiao2022newdesign}
\APACinsertmetastar {%
jiao2022newdesign}%
\begin{APACrefauthors}%
Jiao, R.%
, Luo, J.%
, Malmqvist%
, Johan%
\BCBL {}\ \BBA {} Summers, J.%
\end{APACrefauthors}%
\unskip\
\newblock
\APACrefYearMonthDay{2022}{}{}.
\newblock
{\BBOQ}\APACrefatitle {New Design: Opportunities for Engineering Design in an
  Era of Digital Transformation} {New design: Opportunities for engineering
  design in an era of digital transformation}.{\BBCQ}
\newblock
\APACjournalVolNumPages{Journal of Engineering Design}{xx}{x}{1-5}.
\PrintBackRefs{\CurrentBib}

\bibitem [\protect \citeauthoryear {%
Kopplin%
, Brand%
\BCBL {}\ \BBA {} Reichenberger%
}{%
Kopplin%
\ \protect \BOthers {.}}{%
{\protect \APACyear {2021}}%
}]{%
kopplin2021consumer}
\APACinsertmetastar {%
kopplin2021consumer}%
\begin{APACrefauthors}%
Kopplin, C\BPBI S.%
, Brand, B\BPBI M.%
\BCBL {}\ \BBA {} Reichenberger, Y.%
\end{APACrefauthors}%
\unskip\
\newblock
\APACrefYearMonthDay{2021}{}{}.
\newblock
{\BBOQ}\APACrefatitle {Consumer acceptance of shared e-scooters for urban and
  short-distance mobility} {Consumer acceptance of shared e-scooters for urban
  and short-distance mobility}.{\BBCQ}
\newblock
\APACjournalVolNumPages{Transportation research part D: transport and
  environment}{91}{}{102680}.
\PrintBackRefs{\CurrentBib}

\bibitem [\protect \citeauthoryear {%
Kröyer%
}{%
Kröyer%
}{%
{\protect \APACyear {2021}}%
}]{%
kroyer2021bicycles}
\APACinsertmetastar {%
kroyer2021bicycles}%
\begin{APACrefauthors}%
Kröyer, H.%
\end{APACrefauthors}%
\unskip\
\newblock
\APACrefYearMonthDay{2021}{}{}.
\newblock
{\BBOQ}\APACrefatitle {Bicycles, E-Bikes and Micromobility, A Traffic Safety
  Overview} {Bicycles, e-bikes and micromobility, a traffic safety
  overview}.{\BBCQ}
\newblock
\BIn{} R.~Vickerman\ (\BED), \APACrefbtitle {International Encyclopedia of
  Transportation} {International encyclopedia of transportation}\
  (\BPG~125-138).
\newblock
\APACaddressPublisher{Oxford}{Elsevier}.
\newblock
\begin{APACrefDOI} \doi{https://doi.org/10.1016/B978-0-08-102671-7.10102-2}
  \end{APACrefDOI}
\PrintBackRefs{\CurrentBib}

\bibitem [\protect \citeauthoryear {%
Ling%
, Cherry%
, MacArthur%
\BCBL {}\ \BBA {} Weinert%
}{%
Ling%
\ \protect \BOthers {.}}{%
{\protect \APACyear {2017}}%
}]{%
ling2017differences}
\APACinsertmetastar {%
ling2017differences}%
\begin{APACrefauthors}%
Ling, Z.%
, Cherry, C\BPBI R.%
, MacArthur, J\BPBI H.%
\BCBL {}\ \BBA {} Weinert, J\BPBI X.%
\end{APACrefauthors}%
\unskip\
\newblock
\APACrefYearMonthDay{2017}{}{}.
\newblock
{\BBOQ}\APACrefatitle {Differences of cycling experiences and perceptions
  between e-bike and bicycle users in the United States} {Differences of
  cycling experiences and perceptions between e-bike and bicycle users in the
  united states}.{\BBCQ}
\newblock
\APACjournalVolNumPages{Sustainability}{9}{9}{1662}.
\PrintBackRefs{\CurrentBib}

\bibitem [\protect \citeauthoryear {%
Loustric%
\ \BBA {} Matyas%
}{%
Loustric%
\ \BBA {} Matyas%
}{%
{\protect \APACyear {2020}}%
}]{%
loustric2020}
\APACinsertmetastar {%
loustric2020}%
\begin{APACrefauthors}%
Loustric, I.%
\BCBT {}\ \BBA {} Matyas, M.%
\end{APACrefauthors}%
\unskip\
\newblock
\APACrefYearMonthDay{2020}{}{}.
\newblock
{\BBOQ}\APACrefatitle {Exploring city propensity for the market success of
  micro-electric vehicles} {Exploring city propensity for the market success of
  micro-electric vehicles}.{\BBCQ}
\newblock
\APACjournalVolNumPages{European Transport Research Review}{12}{}{}.
\newblock
\begin{APACrefURL} \url{https://doi.org/10.1186/s12544-020-00416-8}
  \end{APACrefURL}
\newblock
\begin{APACrefDOI} \doi{10.1186/s12544-020-00416-8} \end{APACrefDOI}
\PrintBackRefs{\CurrentBib}

\bibitem [\protect \citeauthoryear {%
MacArthur%
, Cherry%
, Harpool%
\BCBL {}\ \BBA {} Scheppke%
}{%
MacArthur%
\ \protect \BOthers {.}}{%
{\protect \APACyear {2018}}%
}]{%
macarthur2018north}
\APACinsertmetastar {%
macarthur2018north}%
\begin{APACrefauthors}%
MacArthur, J.%
, Cherry, C\BPBI R.%
, Harpool, M.%
\BCBL {}\ \BBA {} Scheppke, D.%
\end{APACrefauthors}%
\unskip\
\newblock
\APACrefYearMonthDay{2018}{}{}.
\newblock
\APACrefbtitle {A North American survey of electric bicycle owners} {A north
  american survey of electric bicycle owners}\ \APACbVolEdTR{}{\BTR{}}.
\newblock
\APACaddressInstitution{Portland, OR 97207}{National Institute for
  Transportation and Communities (NITC)}.
\PrintBackRefs{\CurrentBib}

\bibitem [\protect \citeauthoryear {%
McInnes%
, Healy%
\BCBL {}\ \BBA {} Astels%
}{%
McInnes%
\ \protect \BOthers {.}}{%
{\protect \APACyear {2017}}%
}]{%
mcinnes2017hdbscan}
\APACinsertmetastar {%
mcinnes2017hdbscan}%
\begin{APACrefauthors}%
McInnes, L.%
, Healy, J.%
\BCBL {}\ \BBA {} Astels, S.%
\end{APACrefauthors}%
\unskip\
\newblock
\APACrefYearMonthDay{2017}{}{}.
\newblock
{\BBOQ}\APACrefatitle {hdbscan: Hierarchical density based clustering}
  {hdbscan: Hierarchical density based clustering}.{\BBCQ}
\newblock
\APACjournalVolNumPages{The Journal of Open Source Software}{2}{11}{205}.
\PrintBackRefs{\CurrentBib}

\bibitem [\protect \citeauthoryear {%
{McInnes}%
, {Healy}%
\BCBL {}\ \BBA {} {Melville}%
}{%
{McInnes}%
\ \protect \BOthers {.}}{%
{\protect \APACyear {2018}}%
}]{%
2018arXivUMAP}
\APACinsertmetastar {%
2018arXivUMAP}%
\begin{APACrefauthors}%
{McInnes}, L.%
, {Healy}, J.%
\BCBL {}\ \BBA {} {Melville}, J.%
\end{APACrefauthors}%
\unskip\
\newblock
\APACrefYearMonthDay{2018}{{\APACmonth{02}}}{}.
\newblock
{\BBOQ}\APACrefatitle {{UMAP: Uniform Manifold Approximation and Projection for
  Dimension Reduction}} {{UMAP: Uniform Manifold Approximation and Projection
  for Dimension Reduction}}.{\BBCQ}
\newblock
\APACjournalVolNumPages{ArXiv e-prints}{}{}{}.
\PrintBackRefs{\CurrentBib}

\bibitem [\protect \citeauthoryear {%
{National Association of City Transportation Officials}%
}{%
{National Association of City Transportation Officials}%
}{%
{\protect \APACyear {2020}}%
}]{%
nacto2020}
\APACinsertmetastar {%
nacto2020}%
\begin{APACrefauthors}%
{National Association of City Transportation Officials}.%
\end{APACrefauthors}%
\unskip\
\newblock
\APACrefYearMonthDay{2020}{}{}.
\newblock
\APACrefbtitle {{Shared Micromobility in the U.S.: 2019}} {{Shared
  Micromobility in the U.S.: 2019}}\ \APACbVolEdTR{}{\BTR{}}.
\newblock
\APACaddressInstitution{120 Park Avenue 21st Floor, New York, NY
  10017}{National Association of City Transportation Officials}.
\PrintBackRefs{\CurrentBib}

\bibitem [\protect \citeauthoryear {%
Nguyen%
, Vu%
\BCBL {}\ \BBA {} Nguyen%
}{%
Nguyen%
\ \protect \BOthers {.}}{%
{\protect \APACyear {2020}}%
}]{%
nguyen2020bertweet}
\APACinsertmetastar {%
nguyen2020bertweet}%
\begin{APACrefauthors}%
Nguyen, D\BPBI Q.%
, Vu, T.%
\BCBL {}\ \BBA {} Nguyen, A\BPBI T.%
\end{APACrefauthors}%
\unskip\
\newblock
\APACrefYearMonthDay{2020}{}{}.
\newblock
{\BBOQ}\APACrefatitle {BERTweet: A pre-trained language model for English
  Tweets} {Bertweet: A pre-trained language model for english tweets}.{\BBCQ}
\newblock
\APACjournalVolNumPages{arXiv preprint arXiv:2005.10200}{}{}{}.
\PrintBackRefs{\CurrentBib}

\bibitem [\protect \citeauthoryear {%
Nikitas%
, Tsigdinos%
, Karolemeas%
, Kourmpa%
\BCBL {}\ \BBA {} Bakogiannis%
}{%
Nikitas%
\ \protect \BOthers {.}}{%
{\protect \APACyear {2021}}%
}]{%
nikitas2021cycling}
\APACinsertmetastar {%
nikitas2021cycling}%
\begin{APACrefauthors}%
Nikitas, A.%
, Tsigdinos, S.%
, Karolemeas, C.%
, Kourmpa, E.%
\BCBL {}\ \BBA {} Bakogiannis, E.%
\end{APACrefauthors}%
\unskip\
\newblock
\APACrefYearMonthDay{2021}{}{}.
\newblock
{\BBOQ}\APACrefatitle {Cycling in the era of COVID-19: Lessons learnt and best
  practice policy recommendations for a more bike-centric future} {Cycling in
  the era of covid-19: Lessons learnt and best practice policy recommendations
  for a more bike-centric future}.{\BBCQ}
\newblock
\APACjournalVolNumPages{Sustainability}{13}{9}{4620}.
\PrintBackRefs{\CurrentBib}

\bibitem [\protect \citeauthoryear {%
Pimentel%
\ \BBA {} Lowry%
}{%
Pimentel%
\ \BBA {} Lowry%
}{%
{\protect \APACyear {2020}}%
}]{%
pimentel2020taming}
\APACinsertmetastar {%
pimentel2020taming}%
\begin{APACrefauthors}%
Pimentel, D.%
\BCBT {}\ \BBA {} Lowry, M.%
\end{APACrefauthors}%
\unskip\
\newblock
\APACrefYearMonthDay{2020}{}{}.
\newblock
\APACrefbtitle {Taming and Tapping the Bikeshare Explosion: Review of Shared
  Micro-mobility Laws} {Taming and tapping the bikeshare explosion: Review of
  shared micro-mobility laws}\ \APACbVolEdTR{}{\BTR{}}.
\newblock
\APACaddressInstitution{Seattle, WA}{Pacific Northwest Transportation
  Consortium}.
\PrintBackRefs{\CurrentBib}

\bibitem [\protect \citeauthoryear {%
Price%
, Blackshear%
, Blount~Jr%
\BCBL {}\ \BBA {} Sandt%
}{%
Price%
\ \protect \BOthers {.}}{%
{\protect \APACyear {2021}}%
}]{%
price2021micromobility}
\APACinsertmetastar {%
price2021micromobility}%
\begin{APACrefauthors}%
Price, J.%
, Blackshear, D.%
, Blount~Jr, W.%
\BCBL {}\ \BBA {} Sandt, L.%
\end{APACrefauthors}%
\unskip\
\newblock
\APACrefYearMonthDay{2021}{}{}.
\newblock
{\BBOQ}\APACrefatitle {Micromobility: A Travel Mode Innovation} {Micromobility:
  A travel mode innovation}.{\BBCQ}
\newblock
\APACjournalVolNumPages{Public Roads}{85}{1}{}.
\PrintBackRefs{\CurrentBib}

\bibitem [\protect \citeauthoryear {%
Raptopoulou%
, Basbas%
, Stamatiadis%
\BCBL {}\ \BBA {} Nikiforiadis%
}{%
Raptopoulou%
\ \protect \BOthers {.}}{%
{\protect \APACyear {2020}}%
}]{%
raptopoulou2020first}
\APACinsertmetastar {%
raptopoulou2020first}%
\begin{APACrefauthors}%
Raptopoulou, A.%
, Basbas, S.%
, Stamatiadis, N.%
\BCBL {}\ \BBA {} Nikiforiadis, A.%
\end{APACrefauthors}%
\unskip\
\newblock
\APACrefYearMonthDay{2020}{}{}.
\newblock
{\BBOQ}\APACrefatitle {A first look at e-scooter users} {A first look at
  e-scooter users}.{\BBCQ}
\newblock
\BIn{} \APACrefbtitle {Conference on Sustainable Urban Mobility} {Conference on
  sustainable urban mobility}\ (\BPGS\ 882--891).
\PrintBackRefs{\CurrentBib}

\bibitem [\protect \citeauthoryear {%
Sanh%
, Debut%
, Chaumond%
\BCBL {}\ \BBA {} Wolf%
}{%
Sanh%
\ \protect \BOthers {.}}{%
{\protect \APACyear {2019}}%
}]{%
sanh2019distilbert}
\APACinsertmetastar {%
sanh2019distilbert}%
\begin{APACrefauthors}%
Sanh, V.%
, Debut, L.%
, Chaumond, J.%
\BCBL {}\ \BBA {} Wolf, T.%
\end{APACrefauthors}%
\unskip\
\newblock
\APACrefYearMonthDay{2019}{}{}.
\newblock
{\BBOQ}\APACrefatitle {DistilBERT, a distilled version of BERT: smaller,
  faster, cheaper and lighter} {Distilbert, a distilled version of bert:
  smaller, faster, cheaper and lighter}.{\BBCQ}
\newblock
\APACjournalVolNumPages{arXiv preprint arXiv:1910.01108}{}{}{}.
\PrintBackRefs{\CurrentBib}

\bibitem [\protect \citeauthoryear {%
{Society of Automotive Engineers}%
}{%
{Society of Automotive Engineers}%
}{%
{\protect \APACyear {2019}}%
}]{%
sae}
\APACinsertmetastar {%
sae}%
\begin{APACrefauthors}%
{Society of Automotive Engineers}.%
\end{APACrefauthors}%
\unskip\
\newblock
\APACrefYearMonthDay{2019}{11}{}.
\newblock
\APACrefbtitle {{Taxonomy and Classification of Powered Micromobility
  Vehicles}} {{Taxonomy and Classification of Powered Micromobility Vehicles}}\
  \APACbVolEdTR{}{\BTR{}}.
\newblock
\APACaddressInstitution{400 Commonwealth Drive, Warrendale, PA 15096}{Society
  of Automotive Engineers}.
\PrintBackRefs{\CurrentBib}

\bibitem [\protect \citeauthoryear {%
Sunio%
, Laperal%
\BCBL {}\ \BBA {} Mateo-Babiano%
}{%
Sunio%
\ \protect \BOthers {.}}{%
{\protect \APACyear {2020}}%
}]{%
sunio2020social}
\APACinsertmetastar {%
sunio2020social}%
\begin{APACrefauthors}%
Sunio, V.%
, Laperal, M.%
\BCBL {}\ \BBA {} Mateo-Babiano, I.%
\end{APACrefauthors}%
\unskip\
\newblock
\APACrefYearMonthDay{2020}{}{}.
\newblock
{\BBOQ}\APACrefatitle {Social enterprise as catalyst of transformation in the
  micro-mobility sector} {Social enterprise as catalyst of transformation in
  the micro-mobility sector}.{\BBCQ}
\newblock
\APACjournalVolNumPages{Transportation Research Part A: Policy and
  Practice}{138}{}{145--157}.
\PrintBackRefs{\CurrentBib}

\bibitem [\protect \citeauthoryear {%
Tark%
}{%
Tark%
}{%
{\protect \APACyear {2020}}%
}]{%
tark2020patterns}
\APACinsertmetastar {%
tark2020patterns}%
\begin{APACrefauthors}%
Tark, J.%
\end{APACrefauthors}%
\unskip\
\newblock
\APACrefYearMonthDay{2020}{}{}.
\newblock
\APACrefbtitle {Micromobility Products-Related Deaths, Injuries, and Hazard
  Patterns: 2017–2019} {Micromobility products-related deaths, injuries, and
  hazard patterns: 2017–2019}\ \APACbVolEdTR{}{\BTR{}}.
\newblock
\APACaddressInstitution{4330 East-West Hwy, Bethesda, MD 20814, United
  States}{U.S. Consumer Product Safety Commission}.
\PrintBackRefs{\CurrentBib}

\bibitem [\protect \citeauthoryear {%
Tuncer%
\ \BBA {} Brown%
}{%
Tuncer%
\ \BBA {} Brown%
}{%
{\protect \APACyear {2020}}%
}]{%
tuncer2020scooters}
\APACinsertmetastar {%
tuncer2020scooters}%
\begin{APACrefauthors}%
Tuncer, S.%
\BCBT {}\ \BBA {} Brown, B.%
\end{APACrefauthors}%
\unskip\
\newblock
\APACrefYearMonthDay{2020}{}{}.
\newblock
{\BBOQ}\APACrefatitle {E-scooters on the ground: Lessons for redesigning urban
  micro-mobility} {E-scooters on the ground: Lessons for redesigning urban
  micro-mobility}.{\BBCQ}
\newblock
\BIn{} \APACrefbtitle {Proceedings of the 2020 CHI conference on human factors
  in computing systems} {Proceedings of the 2020 chi conference on human
  factors in computing systems}\ (\BPGS\ 1--14).
\PrintBackRefs{\CurrentBib}

\bibitem [\protect \citeauthoryear {%
Useche%
, Gonzalez-Marin%
, Faus%
\BCBL {}\ \BBA {} Alonso%
}{%
Useche%
\ \protect \BOthers {.}}{%
{\protect \APACyear {2022}}%
}]{%
useche2022environmentally}
\APACinsertmetastar {%
useche2022environmentally}%
\begin{APACrefauthors}%
Useche, S\BPBI A.%
, Gonzalez-Marin, A.%
, Faus, M.%
\BCBL {}\ \BBA {} Alonso, F.%
\end{APACrefauthors}%
\unskip\
\newblock
\APACrefYearMonthDay{2022}{}{}.
\newblock
{\BBOQ}\APACrefatitle {Environmentally friendly, but behaviorally complex? A
  systematic review of e-scooter riders’ psychosocial risk features}
  {Environmentally friendly, but behaviorally complex? a systematic review of
  e-scooter riders’ psychosocial risk features}.{\BBCQ}
\newblock
\APACjournalVolNumPages{PLoS one}{17}{5}{e0268960}.
\PrintBackRefs{\CurrentBib}

\bibitem [\protect \citeauthoryear {%
Vaswani%
\ \protect \BOthers {.}}{%
Vaswani%
\ \protect \BOthers {.}}{%
{\protect \APACyear {2017}}%
}]{%
vaswani2017attention}
\APACinsertmetastar {%
vaswani2017attention}%
\begin{APACrefauthors}%
Vaswani, A.%
, Shazeer, N.%
, Parmar, N.%
, Uszkoreit, J.%
, Jones, L.%
, Gomez, A\BPBI N.%
\BDBL {}Polosukhin, I.%
\end{APACrefauthors}%
\unskip\
\newblock
\APACrefYearMonthDay{2017}{}{}.
\newblock
{\BBOQ}\APACrefatitle {Attention is all you need} {Attention is all you
  need}.{\BBCQ}
\newblock
\APACjournalVolNumPages{Advances in neural information processing
  systems}{30}{}{}.
\PrintBackRefs{\CurrentBib}

\bibitem [\protect \citeauthoryear {%
H.~Yang%
\ \protect \BOthers {.}}{%
H.~Yang%
\ \protect \BOthers {.}}{%
{\protect \APACyear {2020}}%
}]{%
yang2020safety}
\APACinsertmetastar {%
yang2020safety}%
\begin{APACrefauthors}%
Yang, H.%
, Ma, Q.%
, Wang, Z.%
, Cai, Q.%
, Xie, K.%
\BCBL {}\ \BBA {} Yang, D.%
\end{APACrefauthors}%
\unskip\
\newblock
\APACrefYearMonthDay{2020}{}{}.
\newblock
{\BBOQ}\APACrefatitle {Safety of micro-mobility: Analysis of E-Scooter crashes
  by mining news reports} {Safety of micro-mobility: Analysis of e-scooter
  crashes by mining news reports}.{\BBCQ}
\newblock
\APACjournalVolNumPages{Accident Analysis \& Prevention}{143}{}{105608}.
\PrintBackRefs{\CurrentBib}

\bibitem [\protect \citeauthoryear {%
Z.~Yang%
\ \protect \BOthers {.}}{%
Z.~Yang%
\ \protect \BOthers {.}}{%
{\protect \APACyear {2019}}%
}]{%
yang2019xlnet}
\APACinsertmetastar {%
yang2019xlnet}%
\begin{APACrefauthors}%
Yang, Z.%
, Dai, Z.%
, Yang, Y.%
, Carbonell, J.%
, Salakhutdinov, R\BPBI R.%
\BCBL {}\ \BBA {} Le, Q\BPBI V.%
\end{APACrefauthors}%
\unskip\
\newblock
\APACrefYearMonthDay{2019}{}{}.
\newblock
{\BBOQ}\APACrefatitle {Xlnet: Generalized autoregressive pretraining for
  language understanding} {Xlnet: Generalized autoregressive pretraining for
  language understanding}.{\BBCQ}
\newblock
\APACjournalVolNumPages{Advances in neural information processing
  systems}{32}{}{}.
\PrintBackRefs{\CurrentBib}

\bibitem [\protect \citeauthoryear {%
Zagorskas%
\ \BBA {} Burinskien{\.e}%
}{%
Zagorskas%
\ \BBA {} Burinskien{\.e}%
}{%
{\protect \APACyear {2019}}%
}]{%
zagorskas2019challenges}
\APACinsertmetastar {%
zagorskas2019challenges}%
\begin{APACrefauthors}%
Zagorskas, J.%
\BCBT {}\ \BBA {} Burinskien{\.e}, M.%
\end{APACrefauthors}%
\unskip\
\newblock
\APACrefYearMonthDay{2019}{}{}.
\newblock
{\BBOQ}\APACrefatitle {Challenges caused by increased use of E-powered personal
  mobility vehicles in European cities} {Challenges caused by increased use of
  e-powered personal mobility vehicles in european cities}.{\BBCQ}
\newblock
\APACjournalVolNumPages{Sustainability}{12}{1}{273}.
\PrintBackRefs{\CurrentBib}

\bibitem [\protect \citeauthoryear {%
Zhang%
, Yang%
\BCBL {}\ \BBA {} Zhou%
}{%
Zhang%
\ \protect \BOthers {.}}{%
{\protect \APACyear {2022}}%
}]{%
zhang2022disengagement}
\APACinsertmetastar {%
zhang2022disengagement}%
\begin{APACrefauthors}%
Zhang, Y.%
, Yang, X\BPBI J.%
\BCBL {}\ \BBA {} Zhou, F.%
\end{APACrefauthors}%
\unskip\
\newblock
\APACrefYearMonthDay{2022}{}{}.
\newblock
{\BBOQ}\APACrefatitle {Disengagement Cause-and-Effect Relationships Extraction
  Using an NLP Pipeline} {Disengagement cause-and-effect relationships
  extraction using an nlp pipeline}.{\BBCQ}
\newblock
\APACjournalVolNumPages{IEEE Transactions on Intelligent Transportation
  Systems}{}{}{}.
\PrintBackRefs{\CurrentBib}

\bibitem [\protect \citeauthoryear {%
Zhou%
, Ayoub%
, Xu%
\BCBL {}\ \BBA {} Yang%
}{%
Zhou%
\ \protect \BOthers {.}}{%
{\protect \APACyear {2020}}%
}]{%
zhou2020machine}
\APACinsertmetastar {%
zhou2020machine}%
\begin{APACrefauthors}%
Zhou, F.%
, Ayoub, J.%
, Xu, Q.%
\BCBL {}\ \BBA {} Yang, J\BPBI X.%
\end{APACrefauthors}%
\unskip\
\newblock
\APACrefYearMonthDay{2020}{}{}.
\newblock
{\BBOQ}\APACrefatitle {A machine learning approach to customer needs analysis
  for product ecosystems} {A machine learning approach to customer needs
  analysis for product ecosystems}.{\BBCQ}
\newblock
\APACjournalVolNumPages{Journal of Mechanical Design}{142}{1}{}.
\PrintBackRefs{\CurrentBib}

\bibitem [\protect \citeauthoryear {%
Zhou%
, Jiao%
\BCBL {}\ \BBA {} Linsey%
}{%
Zhou%
\ \protect \BOthers {.}}{%
{\protect \APACyear {2015}}%
}]{%
zhou2015latent}
\APACinsertmetastar {%
zhou2015latent}%
\begin{APACrefauthors}%
Zhou, F.%
, Jiao, R.%
\BCBL {}\ \BBA {} Linsey, J\BPBI S.%
\end{APACrefauthors}%
\unskip\
\newblock
\APACrefYearMonthDay{2015}{}{}.
\newblock
{\BBOQ}\APACrefatitle {Latent customer needs elicitation by use case analogical
  reasoning from sentiment analysis of online product reviews} {Latent customer
  needs elicitation by use case analogical reasoning from sentiment analysis of
  online product reviews}.{\BBCQ}
\newblock
\APACjournalVolNumPages{Journal of Mechanical Design}{137}{7}{071401}.
\PrintBackRefs{\CurrentBib}

\end{thebibliography}




\end{document}